\documentclass[conference]{IEEEtran}
\IEEEoverridecommandlockouts
\usepackage{cite}
\usepackage{amsmath,amssymb,amsfonts}
\usepackage{algorithmic}
\usepackage{graphicx}
\usepackage{textcomp}
\def\BibTeX{{\rm B\kern-.05em{\sc i\kern-.025em b}\kern-.08em
    T\kern-.1667em\lower.7ex\hbox{E}\kern-.125emX}}
\usepackage{mwe} 
\usepackage{cite}
\usepackage{amsmath,amssymb,amsfonts}
\usepackage{algorithmic}
\usepackage{graphicx}
\usepackage{textcomp}
\usepackage{lipsum}
\usepackage{float}
\usepackage{fontenc}
\usepackage{setspace}
\usepackage{multirow}
\usepackage{textcomp}
\usepackage{bm}
\usepackage{listings}
\usepackage[mathscr]{euscript}
\usepackage{multirow}
\usepackage{float}
\usepackage[table,xcdraw]{xcolor}
\usepackage{caption}
\usepackage{setspace}
\usepackage{booktabs}
\usepackage{hyperref}
\usepackage{color}
\usepackage{xcolor}
\definecolor{wrong}{rgb}{.8,.349,.1}
\definecolor{right}{rgb}{.3,.7,.1}
\usepackage{amsfonts}

\makeatletter
\newcommand{\linebreakand}{%
  \end{@IEEEauthorhalign}
  \hfill\mbox{}\par
  \mbox{}\hfill\begin{@IEEEauthorhalign}
}
\makeatother

\begin{document}

\title{ICH-SCNet: Intracerebral Hemorrhage Segmentation and Prognosis Classification Network Using CLIP-guided SAM mechanism
}

\author{\IEEEauthorblockN{1\textsuperscript{st} Xinlei Yu$^\dagger$}
\IEEEauthorblockA{\textit{School of Computer Science} \\
\textit{Hangzhou Dianzi University}\\
Hangzhou, China \\
21010237@hdu.edu.cn}\\

\IEEEauthorblockN{4\textsuperscript{th} Hui Jin}
\IEEEauthorblockA{\textit{School of Computer Science} \\
\textit{Hangzhou Dianzi University}\\
Hangzhou, China \\
hui1303101041@gmail.com}\\


\IEEEauthorblockN{7\textsuperscript{th} Qing Wu}
\IEEEauthorblockA{\textit{School of Computer Science} \\
\textit{Hangzhou Dianzi University}\\
Hangzhou, China \\
wuqing@hdu.edu.cn}\\

\and
\IEEEauthorblockN{2\textsuperscript{nd} Ahmed Elazab$^\dagger$}
\IEEEauthorblockA{\textit{School of Biomedical Engineering} \\
\textit{Shenzhen University}\\
Shenzhen, China\\
ahmedelazab@szu.edu.cn}\\

\IEEEauthorblockN{5\textsuperscript{th} Xinchen Jiang}
\IEEEauthorblockA{\textit{School of Management and Economics} \\
\textit{The Chinese University of Hong Kong, Shenzhen}\\
Shenzhen, China \\
122090218@link.cuhk.edu.cn}\\

\IEEEauthorblockN{8\textsuperscript{th} Qinglei Shi}
\IEEEauthorblockA{\textit{School of Medicine} \\
\textit{The Chinese University of Hong Kong, Shenzhen}\\
Shenzhen, China \\
shiqingleicmu@gmail.com}\\
\and
\IEEEauthorblockN{3\textsuperscript{rd} Ruiquan Ge$^*$}
\IEEEauthorblockA{\textit{School of Computer Science} \\
\textit{Hangzhou Dianzi University}\\
Hangzhou, China \\
gespring@hdu.edu.cn}\\

\IEEEauthorblockN{6\textsuperscript{th} Gangyong Jia}
\IEEEauthorblockA{\textit{School of Computer Science} \\
\textit{Hangzhou Dianzi University}\\
Hangzhou, China \\
gangyong@hdu.edu.cn}\\

\IEEEauthorblockN{9\textsuperscript{th} Changmiao Wang$^*$}
\IEEEauthorblockA{\textit{Medical Big Data Lab} \\
\textit{Shenzhen Research Institute of Big Data}\\
Shenzhen, China \\
cmwangalbert@gmail.com}

\thanks{
$\dagger$ These authors have equal contributions.
}
\thanks{$*$ Corresponding authors:  Ruiquan Ge and Changmiao Wang.}
}

\maketitle
\begin{abstract}
Intracerebral hemorrhage (ICH) is the most fatal subtype of stroke and is characterized by a high incidence of disability. Accurate segmentation of the ICH region and prognosis prediction are critically important for developing and refining treatment plans for post-ICH patients. However, existing approaches address these two tasks independently and predominantly focus on imaging data alone, thereby neglecting the intrinsic correlation between the tasks and modalities. This paper introduces a multi-task network, ICH-SCNet, designed for both ICH segmentation and prognosis classification. Specifically, we integrate a SAM-CLIP cross-modal interaction mechanism that combines medical text and segmentation auxiliary information with neuroimaging data to enhance cross-modal feature recognition. Additionally, we develop an effective feature fusion module and a multi-task loss function to improve performance further. Extensive experiments on an ICH dataset reveal that our approach surpasses other state-of-the-art methods. It excels in the overall performance of classification tasks and outperforms competing models in all segmentation task metrics.

\end{abstract}


\section{Introduction}
Intracerebral hemorrhage (ICH) represents a significant public health challenge, affecting over 2 million individuals annually and accounting for approximately 10-15\% of all stroke cases \cite{auy2023intracerebral}. To date, completely effective therapy for all stages of ICH treatment remains elusive \cite{feigin2009worldwide}. Despite advances in medical imaging, particularly computed tomography (CT), and ongoing research on ICH, the rates of mortality and long-term disability in post-ICH patients have not markedly declined \cite{morotti2023intracerebral}. Accurate segmentation of ICH lesions and timely prognosis prediction are indispensable for formulating effective treatment strategies. However, these tasks frequently rely on manual annotations and assessments by neurosurgeons, which are labor-intensive and may be influenced by subjective factors and individual experience \cite{hemphill2015guidelines}. Moreover, most current ICH-related models are neither multifunctional nor fully exploiting non-image modalities \cite{nawabi2021imaging,perez2023deep,ma2023iha}. Such models often overlook the potential for information exchange between different modalities and tasks, which could otherwise enhance overall performance.

\begin{figure*}[t]
    \begin{center}
    \includegraphics[width=0.7\linewidth]{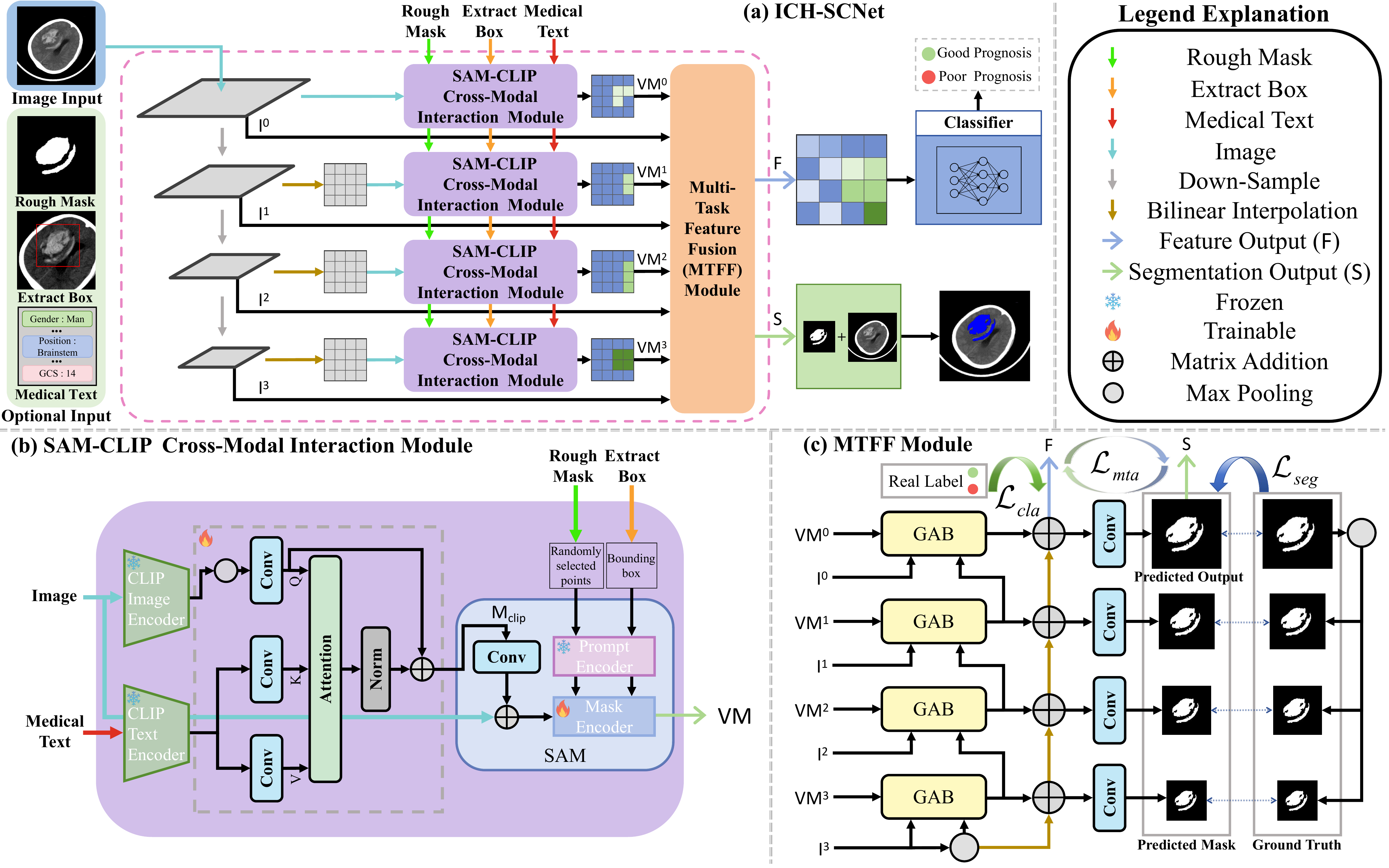}
    \end{center}
   \caption{ (a) Overview of the framework of ICH-SCNet, including depictions of (b) the SAM-CLIP Cross-Modal Interaction module and (c) the MTFF module.}
    \label{overview}
\end{figure*}

 In a medical context, segmentation tasks typically require pixel-level prediction within images, while classification tasks involve the overall categorization of entire images. In this case, segmentation can benefit from the image-level semantic information provided by classification tasks, which assists in accurately delineating lesion areas. Conversely, segmentation tasks offer detailed pixel-level information that aids classification tasks in better understanding the nuances and boundaries within images. Recognizing this interdependence can significantly improve both the precision of segmentation and the reliability of prognosis, ultimately enhancing patient outcomes in clinical practice. Given this symbiotic relationship, some medical networks have embarked on multi-task endeavors \cite{zhu2021dsi,ren2023uncertainty}. However, efforts in multi-task learning for brain regions, specifically for ICH, remain nascent and are sometimes absent. Thus, there exists a practical need for a multi-task network that can concurrently address ICH segmentation and prognosis classification, recognizing and leveraging the inherent interdependence of these tasks.



%

Another significant challenge is the incorporation of diverse and comprehensive medical information into models. Relying solely on imaging data can be inadequate given the complexity and multifaceted nature of medical tasks, which demand a rich variety of information. Particularly in biomedical tasks, especially those involving the brain, it is common to incorporate non-image modal data to guide the model by inputting additional domain knowledge, diagnostic information, and patient demographics. Recent methodologies \cite{nawabi2021imaging,perez2023deep,ma2023iha} have attempted to merge image features with additional medical data through simple concatenation, resulting in limited cross-modal feature interaction. Moreover, evolutionary strategies have endeavored to integrate tabular data \cite{ma2023treatment}, domain-specific knowledge \cite{shan2023gcs}, and textual reports \cite{yu2024ichpro} to achieve a deeper understanding of both image and non-imaging data. However, these approaches still exhibit potential for further refinement, particularly in establishing the internal dependencies of cross-modal features. Recognizing the strengths of both the Segment Anything Model (SAM) \cite{Kirillov2023ICCV} for integrating segmentation auxiliary information, and the Contrastive Language-Image Pre-training (CLIP) \cite{radford2021learning} for cross-modal fusion of texts and images, we propose that the collaboration of these two models can bridge the current gap in merging distinct modalities.




Based on these observations, we innovatively present a SAM-CLIP cross-modal interaction mechanism that incorporates additional medical information to guide segmentation and cross-modal feature extraction. To the best of our knowledge, our ICH-SCNet represents one of the first attempts to develop multi-task models that utilize cross-modal information to concurrently address segmentation and prognosis classification tasks, particularly in the context of ICH. 






\section{Methodology}
As depicted in Fig. \ref{overview} (a), our approach begins with three down-sampling layers that produce image features across four scales, $\{\mathbf{I}^{0},\mathbf{I}^{1},\mathbf{I}^{2},\mathbf{I}^{3}\}$. Due to the use of the standard Vision Transformer (ViT) within the image encoder of the SAM, it is necessary to apply bilinear interpolation to $\{\mathbf{I}^{1},\mathbf{I}^{2},\mathbf{I}^{3}\}$ to achieve appropriate resizing. These interpolated image features, along with clinical text information and extracted boxes and rough masks serving as segmentation prompts, are then input into four distinct SAM-CLIP cross-modal interaction modules. Each module processes the information and yields valid masks at corresponding scales $\{\mathbf{VM}^0,\mathbf{VM}^1,\mathbf{VM}^2,\mathbf{VM}^3\}$.

Subsequently, our model employs a multi-task feature fusion (MTFF) module to integrate the multi-scale image features, which are augmented by the valid masks, through group bridge structures. This integration results in the calculation of the feature output ($\mathbf{F}$), which is subsequently used as input for a DenseNet-121 classifier, as well as the segmentation output ($\mathbf{S}$) that represents the final segmentation result. Moreover, our approach incorporates a multi-task loss function specifically designed to optimize the performance of each task while simultaneously accounting for the inherent connections and correlations among the tasks. This loss function plays a pivotal role in ensuring that the interdependencies between classification and segmentation are effectively captured and leveraged during the model training process.


\subsection{SAM-CLIP Cross-Modal Interaction Module}

As illustrated in Fig. \ref{overview} (b), we initially utilize the CLIP image encoder ($En^{img}_{clip}$) and the CLIP text encoder ($En^{txt}_{clip}$) to process image ($\mathbf{i}$) and text ($\mathbf{t}$) inputs, respectively. We introduce a specialized attention mechanism for adaptive weighting. A max-pooling and convolution layer are applied to the image feature to form Query ($\mathbf{Q}^i$), while two distinct convolution layers process text features, creating Key ($\mathbf{K}^t$) and Value ($\mathbf{V}^t$). An attention mechanism is utilized to facilitate cross-modal interaction, enhanced by residual connections that incorporate $\mathbf{Q}^i$ and the output of a normalization (Norm) layer. The $\mathbf{M}_{clip}$ is computed as follows:
\begin{eqnarray}
    \mathbf{M}_{clip} = Norm\left(Att\left(\mathbf{Q}^i, \mathbf{K}^t, \mathbf{V}^t\right)\right) + En^{img}_{clip}(\mathbf{i}).
\end{eqnarray}

Using the rough mask $\mathbf{M}_{clip}$ from the previous step, we refine it with SAM, guided by the provided bounding box and mask prompts, to generate a valid mask. The bounding box prompt and the rough mask prompt are first synthesized to create a bounding box ($\mathbf{b}$) and a set of randomly selected points ($\mathbf{p}$), which are then processed by the SAM prompt encoder ($En^{pmt}_{sam}$). The final SAM-CLIP valid mask, $\mathbf{mask}_{sam\mbox{-}clip}$, is produced by the SAM mask encoder ($En^{mk}_{sam}$), which refines the rough segmentation with the original image, subject to the constraints imposed by the prompts. The $\mathbf{VM}$ is thus represented as:
\begin{eqnarray}
    \mathbf{VM} = En^{mk}_{sam}\left(\left(Conv(\mathbf{M}_{clip}) + \mathbf{i}\right), En^{pmt}_{sam}(\mathbf{b}, \mathbf{p})\right).
\end{eqnarray}

After investigating various configurations for this module, we opted to employ pre-trained $En^{img}_{clip}$, $En^{txt}_{clip}$, and $En^{pmt}_{sam}$ without modification (frozen). We fine-tune $En^{mk}_{sam}$ and train the attention component, which is enclosed within the gray dashed box, using the ICH dataset to achieve optimal performance.

\subsection{MTFF Module}

Inspired by the group bridge structures in EGE-UNet \cite{ruan2023ege}, we utilize a group aggregation bridge (GAB) to merge the image feature $\mathbf{I}$ in each stage, its corresponding valid mask $\mathbf{VM}$, and the output from the preceding stage. The amalgamated feature from each stage serves as the input for the subsequent stage. By performing pixel-wise addition of these merged features and applying a convolution layer, we generate a predicted mask at each stage. These predictions are then assessed against their respective segmentation labels to compute the loss function, as detailed in Section \ref{loss_sec}. At the final stage, we derive both the segmentation prediction and the multi-scale fused features, denoted as the segmentation output ($\mathbf{S}$) and the feature output ($\mathbf{F}$), respectively.

The GAB is designed with three inputs: the image feature, the valid mask, and the lower-level feature. Initially, these inputs are concatenated sequentially within a group, and four such groups are formed. Subsequently, we apply dilated convolution \cite{yu2016multi} to each group using dilation rates of $\{1,3,5,7\}$ and a kernel size of 3, corresponding to the groups. Finally, the outputs from the four groups are concatenated  to enable interaction between features across different scales, followed by an additional convolution layer to fuse them.

%



\subsection{Loss Function}
\label{loss_sec}
We meticulously designed a loss function for ICH-SCNet to facilitate optimization. This comprehensive loss function encompasses individual losses for both segmentation and classification tasks, as well as a multi-task aware loss to establish and exploit correlations between the tasks.

\textbf{Segmentation Loss. } For each scale level, we calculate the Dice Similarity Coefficient (DSC) and Jaccard index (Jaccard), as introduced in Section \ref{metric}, to evaluate the congruence between the predicted masks and the ground truths. The overall segmentation loss $\mathscr{L}_{seg}$ is formulated as a weighted sum of these metrics across the four scales, with weights assigned as $1, 0.75, 0.5, 0.25$ for scales from $0$ to $3$, respectively, from top to bottom. Recognizing the necessity for masks at varying scales, we have implemented deep supervision \cite{zhou2019unet++}. The segmentation loss can be expressed as follows:
\begin{eqnarray}
    \mathscr{L}_{seg} = \sum_{i=0}^{3} \gamma_i \times \left(\operatorname{DSC}(s_i, \hat{s_i}) + \operatorname{Jaccard}(s_i, \hat{s_i})\right),
\end{eqnarray}
where DSC and Jaccard are the segmentation metrics as defined in Section \ref{metric}, $\hat{s}$ is the segmentation prediction, and $s$ is the ground truth. The weight $\gamma_i$ for different scales of segmentation loss is set to $1, 0.75, 0.5, 0.25$ from $i=0$ to $3$ in our work.

\textbf{Classification Loss.} For prognosis classification, we employ a weighted cross-entropy loss to quantify the distance between the predicted outcomes and the true labels. Mathematically, the classification loss $\mathscr{L}_{cla}$ is expressed as follows:
\begin{eqnarray}
    \mathscr{L}_{cla} = -\xi_{1} \cdot \left(\mathbb{I}\{l=1\} \log \hat{p}_{1}\right) - \xi_{0} \cdot \left(\mathbb{I}\{l=0\} \log \hat{p}_{0}\right),
\end{eqnarray}
where $\mathbb{I}\{\cdot\}$ signifies the indicator function, $\hat{p}$ represents the predicted probability, and $l$ is the true label of the prognosis. The term $\xi$ refers to the class-specific weights derived from the inverse frequency of the two classes within the entire dataset, thereby mitigating the effects of potential sample imbalance.

%


\textbf{Multi-task Aware Loss.}  Relying solely on $\mathscr{L}_{seg}$ or $\mathscr{L}_{cla}$ is insufficient, as this approach may result in an optimization process excessively biased towards one task to the detriment of the other \cite{zhu2021dsi}. In our approach, we direct the feature output $\mathbf{F}$ through a softmax layer to obtain pixel-level probability values, represented by $\mathbf{P}$. Subsequently, we employ the Jensen-Shannon (JS) Divergence to quantify the discrepancy between $\mathbf{P}$ and the segmentation output $\mathbf{S}$. This divergence is expressed as $\mathscr{L}_{mta}$ and serves to ensure and achieve the desired consistency between the segmentation and classification tasks. The divergence can be presented as follows:
\begin{eqnarray}
    \mathscr{L}_{mta} &=& D_{JS}(P \| S) = D_{KL}(P \| S) + D_{KL}(S \| P),
\end{eqnarray}
where $D_{KL}(\cdot \|\cdot)$ represents the Kullback-Leibler (KL) Divergence and $N$ is the number of training samples in a batch.



\textbf{Total Loss.} 
The total loss of ICH-SCNet can be expressed as follows:
\begin{eqnarray}
    \mathscr{L}_{total} = \mathscr{L}_{mta} + \alpha \mathscr{L}_{seg} + \beta \mathscr{L}_{cla},
\end{eqnarray}
where $\alpha$ and $\beta$ are weighting hyperparameters that harmonize the contributions of $\mathscr{L}_{seg}$ and $\mathscr{L}_{cla}$, respectively. These hyperparameters are experimentally determined, with values set to 0.2 for $\alpha$ and 0.8 for $\beta$, to ensure an appropriate balance between segmentation and classification objectives.

\begin{table*}[t] 
\centering
\caption{Comparison of single- and multi-task.}
\setlength{\tabcolsep}{0.9mm}{
\resizebox{0.68\linewidth}{!}{
\begin{tabular}{c|cccc|cccc}
\hline
\multicolumn{1}{c|}{\multirow{2}{*}{Method}} & \multicolumn{4}{c|}{Classification}  & \multicolumn{4}{c}{Segmentation} \\ 
\cline{2-9} \multicolumn{1}{c|}{} & \multicolumn{1}{l}{Acc$\uparrow$(\%)} & \multicolumn{1}{l}{Rec$\uparrow$(\%)} & \multicolumn{1}{l}{Pre$\uparrow$(\%)} & \multicolumn{1}{l|}{AUC$\uparrow$($10^{-2}$)} & DSC$\uparrow$(\%) & Jaccard$\uparrow$(\%) & 95HD$\downarrow$ 
& PRO$\uparrow$(\%) \\ \hline
Cla-only& 85.10 & 79.52 & 87.47 & 90.74 & \textbf{-} & \textbf{-} & \textbf{-} & \textbf{-}\\
Seg-only & \textbf{-} & \textbf{-} & \textbf{-} & \textbf{-} & 87.00 & 75.28 & 5.42 & 83.56 \\ \hline
\textbf{ICH-SCNet (cla+seg)} & \textbf{90.71} & \textbf{84.99}&\textbf{91.91}&\textbf{95.11}&\textbf{88.32}&\textbf{76.70}&\textbf{2.77}&\textbf{86.43} \\ \hline
\end{tabular}}}
\label{taskablation}
\end{table*}

\begin{table*}[t] 
\centering
\caption{Ablation studies of the key components of our model.}
\setlength{\tabcolsep}{0.9mm}{
\resizebox{0.68\linewidth}{!}{
\begin{tabular}{c|cccc|cccc}
\hline
\multicolumn{1}{c|}{\multirow{2}{*}{Method}} & \multicolumn{4}{c|}{Classification}  & \multicolumn{4}{c}{Segmentation} \\ 
\cline{2-9} \multicolumn{1}{c|}{} & \multicolumn{1}{l}{Acc$\uparrow$(\%)} & \multicolumn{1}{l}{Rec$\uparrow$(\%)} & \multicolumn{1}{l}{Pre$\uparrow$(\%)} & \multicolumn{1}{l|}{AUC$\uparrow$($10^{-2}$)} & DSC$\uparrow$(\%) & Jaccard$\uparrow$(\%) & 95HD$\downarrow$ 
& PRO$\uparrow$(\%) \\ \hline
ICH-SCNet (SAM) & \textbf{-} & \textbf{-} & \textbf{-} & \textbf{-} & 84.61 & 66.31 & 13.33 & 79.62 \\
ICH-SCNet (CLIP) & 84.22 & 80.88 & 86.98 & 89.97 & \textbf{-} & \textbf{-} & \textbf{-} & \textbf{-} \\ 
ICH-SCNet (SAM-CLIP) & \underline{88.73} & \underline{83.21} & \underline{89.44} & 91.48 & 85.26 & \underline{72.27} & \underline{4.09} &  82.12\\ 
ICH-SCNet (SAM+MTFF) & \textbf{-} & \textbf{-} & \textbf{-} & \textbf{-} & \underline{86.94} & 67.92 & 8.87 & \underline{82.22} \\ 
ICH-SCNet (CLIP+MTFF) & 87.53 & 82.97 & 86.50 & \underline{91.80} & \textbf{-} & \textbf{-} & \textbf{-} & 
\textbf{-} \\  \hline
\textbf{ICH-SCNet(Ours)} & \textbf{90.71} & \textbf{84.99}&\textbf{91.91}&\textbf{95.11}&\textbf{88.32}&\textbf{76.70}&\textbf{2.77}&\textbf{86.43} \\ \hline
\end{tabular}}}
\label{modelablation}
\end{table*}

\section{Experiments}
\subsection{Experimental Setup}
\label{metric}
\textbf{Dataset. }Our study utilized a proprietary ICH dataset provided by our collaborating hospital, comprising data from 294 patients and a total of 1,907 effective CT slices. The dataset includes 865 slices indicating good prognoses and 1,042 slices suggestive of poor prognoses. Each patient record in the dataset encompasses a single CT image, associated medical text, and a prognosis label, which is adjudicated using the Glasgow Outcome Scale (GOS) \cite{mcmillan2016glasgow}. A GOS score of $\geq 4$ indicates a good prognosis, while a score of $<4$ indicates a poor prognosis. Additionally, the dataset provides ground truth segmentation, bounding box annotations, and approximate masks for the hemorrhage regions. The medical text accompanying each case contains critical information, such as the patient's age, gender, length of hospital stay, time from onset to the CT scan, Glasgow Coma Scale score \cite{teasdale2014glasgow}, treatment methods, and the location and volume of the hemorrhage. 



\textbf{Implementation Details. } In our experimental setup, we utilized the last transformer layer of the pre-trained ViT-L model as CLIP encoders, while the ViT-H pre-trained model was employed for the SAM. During the training phase, we adopted the AdamW optimizer \cite{loshchilov2018decoupled} with an initial learning rate set to $1\times10^{-4}$ and a batch size of 32. The training was conducted throughout 25 epochs, focusing only on the parts of the model that were amenable to training. The computational experiments were performed on high-performance servers equipped with 4 NVIDIA A100 GPUs, ensuring substantial processing power. Moreover, all experiments were conducted using five-fold cross-validation to fully utilize our dataset and ensure robust evaluation.



\subsection{Experimental Results}
\textbf{Effectiveness of Multi-task Network.} 

As presented in Table \ref{taskablation}, a comparative analysis revealed that the performance of the network when operating on segmentation and classification tasks in isolation is inferior to that of our multi-task framework. Furthermore, it is noteworthy that the classification task benefits more significantly from the multi-task setup than the segmentation task. Specifically, the segmentation metric, DSC shows an improvement of 1.32\%, whereas the classification Acc experiences a substantial gain of 5.61\% when comparing multi-tasking to single-tasking scenarios. 


\begin{table*}[t] 
\centering
\caption{Comparisons with state-of-art (SOTA) methods.}
\setlength{\tabcolsep}{0.9mm}{
\resizebox{0.68\linewidth}{!}{
\begin{tabular}{c|c|llll|llll}
\hline
\multicolumn{1}{c|}{\multirow{2}{*}{No.}} & \multicolumn{1}{c|}{\multirow{2}{*}{Method}} & \multicolumn{4}{c|}{Classification}  & \multicolumn{4}{c}{Segmentation} \\ 
\cline{3-10} &\multicolumn{1}{|c|}{} & \multicolumn{1}{l}{Acc$\uparrow$(\%)} & \multicolumn{1}{l}{Rec$\uparrow$(\%)} & \multicolumn{1}{l}{Pre$\uparrow$(\%)} & \multicolumn{1}{l|}{AUC$\uparrow$($10^{-2}$)} & DSC$\uparrow$(\%) & Jaccard$\uparrow$(\%) & 95HD$\downarrow$ 
& PRO$\uparrow$(\%) \\ \hline
1&Image-based\cite{nawabi2021imaging} & 81.51 & 76.20 & 82.85 & 80.79 & \textbf{\quad-} & \textbf{\quad-} & \textbf{\quad-} & \textbf{\quad-}\\
2&DL-PP\cite{perez2023deep} & 83.32 & 79.68 & 83.95 & 80.74
& \textbf{\quad-} & \textbf{\quad-} & \textbf{\quad-} & \textbf{\quad-} \\
3&GCS-ICHNet\cite{shan2023gcs} & 85.08 & 81.88 & 87.25 & 85.90 & \textbf{\quad-} & \textbf{\quad-} & \textbf{\quad-} & \textbf{\quad-} \\
4&TOP-GPM\cite{ma2023treatment} & \underline{89.34} & 82.54 & \underline{92.38} & 84.61 & \textbf{\quad-} & \textbf{\quad-} & \textbf{\quad-} & \textbf{\quad-} \\
5&ICHPro\cite{yu2024ichpro}& 89.11 & \underline{84.56} & 91.02 & \underline{94.29}
& \textbf{\quad-} & \textbf{\quad-} & \textbf{\quad-} & \textbf{\quad-} \\ \hline
6&AMCons\cite{silva2022constrained} & \textbf{\quad-} & \textbf{\quad-} & \textbf{\quad-} & \textbf{\quad-} & \underline{85.18} & 72.27 & 4.71 & 81.88 \\ 
7&IHA-Net\cite{ma2023iha} & \textbf{\quad-} & \textbf{\quad-} & \textbf{\quad-} & \textbf{\quad-} & 84.63 & 71.70 & 4.23 & 80.79 \\ 
8&nnUNet\cite{isensee2021nnu}& \textbf{\quad-} & \textbf{\quad-} & \textbf{\quad-} & \textbf{\quad-} & 83.84 & 70.73 & 10.62 & 80.19  \\
9&SAM\cite{Kirillov2023ICCV} & \textbf{\quad-} & \textbf{\quad-} & \textbf{\quad-} & \textbf{\quad-} & 84.99 & 71.68 & 5.42 & 78.55 \\ 
10&MedSAM\cite{ma2024segment} & \textbf{\quad-} & \textbf{\quad-} & \textbf{\quad-} & \textbf{\quad-} & 85.14 & \underline{72.89} & \underline{4.20} & \underline{82.68} \\ \hline
11&ResGANet\cite{cheng2022resganet} & 82.61 & 75.13 & 83.06 & 83.94 & 81.94 & 67.54 & 12.42 & 82.17 \\ 
12&SelfMedMAE\cite{zhou2023self} & 84.84 & 79.99 & 88.02 & 85.76 &  82.36 & 67.06 & 10.68 & 82.06\\ \hline
\textbf{$\star$} & \textbf{ICH-SCNet(Ours)} & \textbf{90.71}{\color{right}\textbf{\scriptsize{$\uparrow$}1.37}} & \textbf{84.99}{\color{right}\textbf{\scriptsize{$\uparrow$}0.43}}&\textbf{91.91}{\color{wrong}\textbf{\scriptsize{$\downarrow$}0.47}}&\textbf{95.11}
{\color{right}\textbf{\scriptsize{$\uparrow$}0.82}}
&\textbf{88.32}{\color{right}\textbf{\scriptsize{$\uparrow$}3.14}}
&\textbf{76.70}{\color{right}\textbf{\scriptsize{$\uparrow$}3.81}}
&\textbf{2.77}{\color{right}\textbf{\scriptsize{$\downarrow$}1.43}}
&\textbf{86.43}{\color{right}\textbf{\scriptsize{$\uparrow$}3.75}} \\ \hline
\end{tabular}}}
\label{comparison}
\end{table*}



\begin{figure*}[t]
    \begin{center}
    \includegraphics[width=0.68\linewidth]{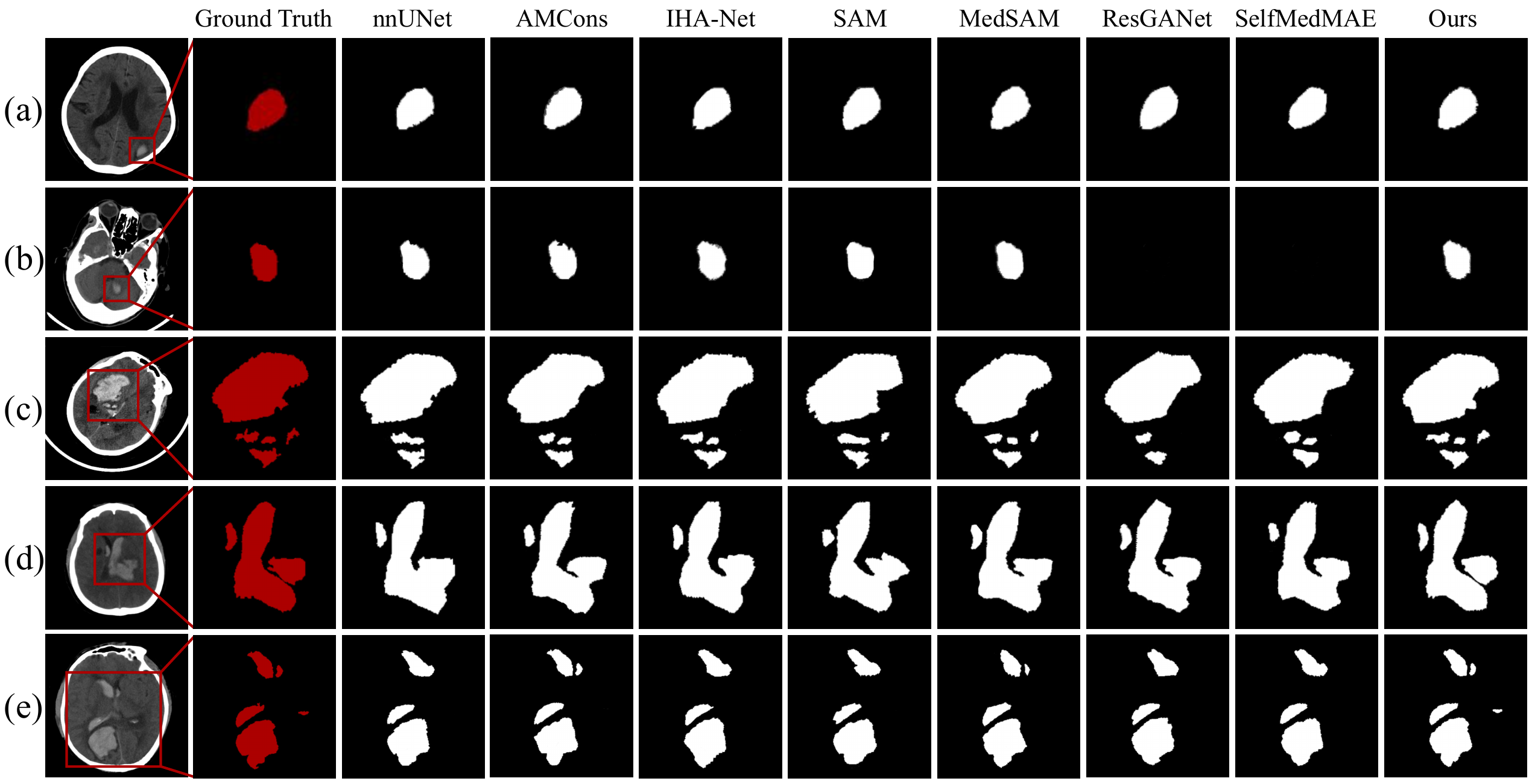}
    \end{center}
   \caption{ Comparisons from segmentation results of SOTA methods and our ICH-SCNet.}
    \label{visual}
\end{figure*}

\textbf{Ablation Study. }Table \ref{modelablation} presents an ablation study designed to ascertain the contribution of key components within our model, particularly focusing on the SAM-CLIP cross-modal interaction module and the MTFF module. Our investigation involved replacing the SAM-CLIP module with either SAM or CLIP and contrasting the MTFF module by directly concatenating all features. In the experiments, we first replaced the SAM-CLIP interaction module with independent SAM or CLIP modules while maintaining the multi-scale structure, which correspondingly reduced the input modalities. Next, we replaced the MTFF module with a direct concatenation structure to verify its effectiveness.

The results demonstrate that the synergistic integration of SAM and CLIP leads to a notable improvement, with Acc increasing by 4.51\% and the DSC by 0.65 compared to the performance of SAM or CLIP independently. This indicates that SAM is adept at providing accurate baseline segmentation by capturing low-level visual details, benefiting from its capacity for broad segmentation and generalization. On the other hand, CLIP enhances the model's representational capabilities by fusing multimodal information from text and images, thereby capturing high-level semantic information across the entire image. Furthermore, the ablation of the MTFF module results in a substantial decline in performance, with accuracy dropping by 1.98\% and DSC by 3.06, thus proving the importance of the MTFF module for effective multi-scale feature fusion. This confirms that the MTFF block plays a pivotal role in harmonizing features across different scales and modalities, which is essential for achieving superior performance in our multi-task model.


\textbf{Comparison with SOTA Methods. }We further evaluated ICH-SCNet against various SOTA methods in the domains of ICH segmentation and prognosis outcome classification. First, we compared our network with five classification models, including two single-modal approaches (No.1-2) and three cross-modal methods that incorporate additional medical information (No.3-5). For segmentation, our comparisons included two specialized ICH networks (No.6-7) and three general medical segmentation networks (No.8-10). Due to the scarcity of multi-task methods addressing both segmentation and prognosis classification for ICH, we also compared our method to two generic medical multi-task networks (No.11-12).


Table \ref{comparison} presents the comparative results from various SOTA methodologies. Our proposed ICH-SCNet demonstrated substantial improvements in model performance, as evidenced by assessments across four segmentation metrics and four classification metrics. For classification, our method leads in three metrics and is only marginally lower than TOP-GPM in Pre. Compared to single-task classification networks that rely exclusively on imaging, ICH-SCNet holds a considerable advantage. Even when juxtaposed with other cross-modal approaches, the benefits conferred by multi-task information sharing are sustained. 

In segmentation tasks, our method outperforms competing approaches comprehensively. Even when compared to two SAM-based approaches, ICH-SCNet secures superior performance across all four metrics. The two general multi-task methods display limited competitiveness, which can be attributed to their deficiency in mechanisms for effective task coordination. Overall, our ICH-SCNet improved performance by 1.27\% in Acc, 0.43\% in Rec, and 0.82 in the AUC for classification. For segmentation, the improvements were 3.14\% in DSC, 3.81\% in Jaccard, 1.43 in 95HD, and 3.75\% in PRO.

To vividly illustrate the segmentation efficacy, we conducted visualization experiments, the results of which are depicted in Fig. \ref{visual}. In straightforward segmentation scenarios, such as those depicted in (a), all methods achieve satisfactory segmentation accuracy. In contrast, for more complex imagery, ICH-SCNet displays superior competitiveness and segmentation precision. For example, clinically challenging scenarios such as multiple bleeding points (c), (e), small bleeding volumes (b), and indistinct boundaries (d) highlight the proficiency of our method.

\section{Conclusion}
This paper presents a novel multi-task network for ICH segmentation and prognosis outcome classification by employing a SAM-CLIP cross-modal interaction mechanism. The network adeptly integrates multi-scale imaging with pertinent medical text and leverages segmentation auxiliary data, guided by a designated multi-task loss function. Through rigorous and exhaustive experimental evaluation on an in-house dataset, the efficacy of our model has been verified.

Regarding constraints, future research will endeavor to amplify the trainable aspects of the network while concurrently diminishing its parameter count and computational complexity. Addressing the negative impact of missing modalities also remains a key issue we are striving to overcome. Furthermore, we posit that this framework holds promise for applicability across a broader spectrum of medical multi-task challenges.

\section{ACKNOWLEDGMENTS}
This work was supported by the Open Project Program of the State Key Laboratory of CAD\&CG, Zhejiang University (No. A2410), Zhejiang Provincial Natural Science Foundation of China (No. LY21F020017), National Natural Science Foundation of China (No. 61702146, 62076084), GuangDong Basic and Applied Basic Research Foundation  (No. 2022A1515110570), Guangxi Key R\&D Project  (No. AB24010167), the Project (No. 20232ABC03A25), Innovation Teams of Youth Innovation in Science, Shenzhen Stability Science Program 2022 (No. 2023SC0073), Technology of High Education Institutions of Shandong Province (No. 2021KJ088), Shenzhen Science and Technology Program (No. KCXFZ20201221173008022), Longgang District Medical and Health Technology Tackling Project (No. LGKCYLWS2023018). 
%
%

\bibliographystyle{splncs04}
\bibliography{conference_101719}

\begin{thebibliography}{10}
\providecommand{\url}[1]{\texttt{#1}}
\providecommand{\urlprefix}{URL }
\providecommand{\doi}[1]{https://doi.org/#1}

\bibitem{perez2023deep}
Perez~del Barrio, A., Esteve~Dom{\'\i}nguez, A.S., Men{\'e}ndez Fern{\'a}ndez-Miranda, P., Sanz~Bell{\'o}n, P., Rodr{\'\i}guez~Gonz{\'a}lez, D., Lloret~Iglesias, L., Marques~Fraguela, E., Gonz{\'a}lez~Mandly, A.A., Vega, J.A.: A deep learning model for prognosis prediction after intracranial hemorrhage. Journal of Neuroimaging  \textbf{33}(2),  218--226 (2023)

\bibitem{cheng2022resganet}
Cheng, J., Tian, S., Yu, L., Gao, C., Kang, X., Ma, X., Wu, W., Liu, S., Lu, H.: Resganet: Residual group attention network for medical image classification and segmentation. Medical Image Analysis  \textbf{76},  102313 (2022)

\bibitem{feigin2009worldwide}
Feigin, V.L., Lawes, C.M., Bennett, D.A., Barker-Collo, S.L., Parag, V.: Worldwide stroke incidence and early case fatality reported in 56 population-based studies: a systematic review. The Lancet Neurology  \textbf{8}(4),  355--369 (2009)

\bibitem{hemphill2015guidelines}
Hemphill~III, J.C., Greenberg, S.M., Anderson, C.S., Becker, K., Bendok, B.R., Cushman, M., Fung, G.L., Goldstein, J.N., Macdonald, R.L., Mitchell, P.H., et~al.: Guidelines for the management of spontaneous intracerebral hemorrhage: a guideline for healthcare professionals from the american heart association/american stroke association. Stroke  \textbf{46}(7),  2032--2060 (2015)

\bibitem{isensee2021nnu}
Isensee, F., Jaeger, P.F., Kohl, S.A., Petersen, J., Maier-Hein, K.H.: nnu-net: a self-configuring method for deep learning-based biomedical image segmentation. Nature Methods  \textbf{18}(2),  203--211 (2021)

\bibitem{Kirillov2023ICCV}
Kirillov, A., Mintun, E., Ravi, N., Mao, H., Rolland, C., Gustafson, L., Xiao, T., Whitehead, S., Berg, A.C., Lo, W.Y., Dollar, P., Girshick, R.: Segment anything. In: Proceedings of the IEEE/CVF International Conference on Computer Vision (ICCV). pp. 4015--4026 (2023)

\bibitem{loshchilov2018decoupled}
Loshchilov, I., Hutter, F.: Decoupled weight decay regularization. In: International Conference on Learning Representations (ICLR) (2019)

\bibitem{ma2024segment}
Ma, J., He, Y., Li, F., Han, L., You, C., Wang, B.: Segment anything in medical images. Nature Communications  \textbf{15}(1), ~654 (2024)

\bibitem{ma2023treatment}
Ma, W., Chen, C., Abrigo, J., Mak, C.H.K., Gong, Y., Chan, N.Y., Han, C., Liu, Z., Dou, Q.: Treatment outcome prediction for intracerebral hemorrhage via generative prognostic model with imaging and tabular data. In: International Conference on Medical Image Computing and Computer-Assisted Intervention (MICCAI). pp. 715--725. Springer (2023)

\bibitem{ma2023iha}
Ma, Y., Ren, F., Li, W., Yu, N., Zhang, D., Li, Y., Ke, M.: Iha-net: An automatic segmentation framework for computer-tomography of tiny intracerebral hemorrhage based on improved attention u-net. Biomedical Signal Processing and Control  \textbf{80},  104320 (2023)

\bibitem{mcmillan2016glasgow}
McMillan, T., Wilson, L., Ponsford, J., Levin, H., Teasdale, G., Bond, M.: The glasgow outcome scale—40 years of application and refinement. Nature Reviews Neurology  \textbf{12}(8),  477--485 (2016)

\bibitem{morotti2023intracerebral}
Morotti, A., Boulouis, G., Dowlatshahi, D., Li, Q., Shamy, M., Salman, R.A.S., Rosand, J., Cordonnier, C., Goldstein, J.N., Charidimou, A.: Intracerebral haemorrhage expansion: definitions, predictors, and prevention. The Lancet Neurology  \textbf{22}(2),  159--171 (2023)

\bibitem{nawabi2021imaging}
Nawabi, J., Kniep, H., Elsayed, S., Friedrich, C., Sporns, P., Rusche, T., B{\"o}hmer, M., Morotti, A., Schlunk, F., D{\"u}hrsen, L., et~al.: Imaging-based outcome prediction of acute intracerebral hemorrhage. Translational Stroke Research  \textbf{12},  958--967 (2021)

\bibitem{auy2023intracerebral}
Puy, L., Parry-Jones, A.R., Sandset, E.C., Dowlatshahi, D., Ziai, W., Cordonnier, C.: Intracerebral haemorrhage. Nature Reviews Disease Primers  \textbf{9}(1), ~14 (2023)

\bibitem{radford2021learning}
Radford, A., Kim, J.W., Hallacy, C., Ramesh, A., Goh, G., Agarwal, S., Sastry, G., Askell, A., Mishkin, P., Clark, J., et~al.: Learning transferable visual models from natural language supervision. In: International Conference on Machine Learning (ICML). pp. 8748--8763. PMLR (2021)

\bibitem{ren2023uncertainty}
Ren, K., Zou, K., Liu, X., Chen, Y., Yuan, X., Shen, X., Wang, M., Fu, H.: Uncertainty-informed mutual learning for joint medical image classification and segmentation. In: International Conference on Medical Image Computing and Computer-Assisted Intervention (MICCAI). pp. 35--45. Springer (2023)

\bibitem{ruan2023ege}
Ruan, J., Xie, M., Gao, J., Liu, T., Fu, Y.: Ege-unet: an efficient group enhanced unet for skin lesion segmentation. In: International Conference on Medical Image Computing and Computer-Assisted Intervention (MICCAI). pp. 481--490. Springer (2023)

\bibitem{shan2023gcs}
Shan, X., Li, X., Ge, R., Wu, S., Elazab, A., Zhu, J., Zhang, L., Jia, G., Xiao, Q., Wan, X., et~al.: Gcs-ichnet: Assessment of intracerebral hemorrhage prognosis using self-attention with domain knowledge integration. In: International Conference on Bioinformatics and Biomedicine (BIBM). pp. 2217--2222. IEEE (2023)

\bibitem{silva2022constrained}
Silva-Rodr{\'\i}guez, J., Naranjo, V., Dolz, J.: Constrained unsupervised anomaly segmentation. Medical Image Analysis  \textbf{80},  102526 (2022)

\bibitem{teasdale2014glasgow}
Teasdale, G., Maas, A., Lecky, F., et~al.: The glasgow coma scale at 40 years: standing the test of time. The Lancet Neurology  \textbf{13}(8),  844--854 (2014)

\bibitem{yu2016multi}
Yu, F., Koltun, V.: Multi-scale context aggregation by dilated convolutions. In: International Conference on Learning Representations (ICLR) (2016)

\bibitem{yu2024ichpro}
Yu, X., Li, X., Ge, R., Wu, S., Elazab, A., Zhu, J., Zhang, L., Jia, G., Xu, T., Wan, X., Wang, C.: Ichpro: Intracerebral hemorrhage prognosis classification via joint-attention fusion-based 3d cross-modal network. In: 2024 IEEE International Symposium on Biomedical Imaging (ISBI). pp.~1--5 (2024)

\bibitem{zhou2023self}
Zhou, L., Liu, H., Bae, J., He, J., Samaras, D., Prasanna, P.: Self pre-training with masked autoencoders for medical image classification and segmentation. In: International Symposium on Biomedical Imaging (ISBI). pp.~1--6. IEEE (2023)

\bibitem{zhou2019unet++}
Zhou, Z., Siddiquee, M.M.R., Tajbakhsh, N., Liang, J.: Unet++: Redesigning skip connections to exploit multiscale features in image segmentation. IEEE Transactions on Medical Imaging (TMI)  \textbf{39}(6),  1856--1867 (2019)

\bibitem{zhu2021dsi}
Zhu, M., Chen, Z., Yuan, Y.: Dsi-net: Deep synergistic interaction network for joint classification and segmentation with endoscope images. IEEE Transactions on Medical Imaging (TMI)  \textbf{40}(12),  3315--3325 (2021)

\end{thebibliography}

\end{document}